# PeCoQ: A Dataset for Persian Complex Question Answering over Knowledge Graph

Romina Etezadi
*Natural Language Processing Lab*
*Shahid Beheshti University*
Tehran, Iran
ro.etezadi@mail.sbu.ac.ir

Mehrnoush Shamsfard
*Natural Language Processing Lab*
*Shahid Beheshti University*
Tehran, Iran
m-shams@sbu.ac.ir

*Abstract*—Question answering systems may find the answers to users' questions from either unstructured texts or structured data such as knowledge graphs. Answering questions using supervised learning approaches including deep learning models need large training datasets. In recent years, some datasets have been presented for the task of Question answering over knowledge graphs, which is the focus of this paper. Although many datasets in English were proposed, there have been a few question answering datasets in Persian. This paper introduces *PeCoQ*, a dataset for Persian question answering. This dataset contains 10,000 complex questions and answers extracted from the Persian knowledge graph, FarsBase. For each question, the SPARQL query and two paraphrases that were written by linguists are provided as well. There are different types of complexities in the dataset, such as multi-relation, multi-entity, ordinal, and temporal constraints. In this paper, we discuss the dataset's characteristics and describe our methodology for building it.

## I. Introduction

Over the last years, one of the most popular and still open and challenging tasks in natural language processing is answering users' questions. QA systems can be categorized into closed-domain and open-domain systems. A closed-domain system is capable of answering questions that are in a specific domain such as health care, tourism, etc. On the other hand, an open-domain system can answer questions about everything. In past decades knowledge graphs like Freebase [5] and DBpedia [2] have played an important role in answering open-domain questions as they store a lot of information in linked data structure. The focus of question answering systems using knowledge graphs (KGQA) is more on translating a natural language question into a formal language such as $\lambda$-DCS [11] or SPARQL. To achieve this goal, one can use either rule-based approaches [10] or machine learning techniques, which have attracted researchers' attention recently.

Many systems are introduced that achieved acceptable results on simple questions [8] which contaire only a single fact with the form of $<subject, relation, object>$, but handling questions that need finding more than one fact (complex questions) still remains a challenge in question answering task.

The main part of machine learning approaches is the need for a large dataset for training. Therefore, having a related dataset can lead to an acceptable and accurate result for solving any task. In case of question answering over knowledge graphs, there are many datasets in English such as WebQuestions [4] and SimpleQuestions [6] for simple questions and ComplexQuestions [3], ComplexWebQuestions [13], and LC-Quad [9] for complex questions. There is a closed-domain dataset in Persian, Rasayel&massayel, that has both simple and complex questions [7]. To the best of our knowledge, there are no previous datasets for open domain question answering in the Persian language, especially for complex questions. This work aims to present a dataset in Persian that consists of 10,000 complex questions over FarsBase [12] a Persian language knowledge graph. PeCoQ can be used for systems that try to answer complex questions using KG and also systems try to convert complex questions to logical forms. The main contributions of this paper are:

- A large dataset with 10,000 complex questions with their corresponding SPARQL queries over FarsBase and their answers.
- All questions contain at least two paraphrases (written by linguists) besides the machine-generated ones.
- In addition to SPARQL queries, entities mentioned in the question and their relations are provided as well.
- There is a good variety of complexity in the dataset such as multi-relation, multi-entity, comparative, superlative, aggregation, and temporal questions which we will discuss in section IV.

The rest of this paper is as follows: Section 2 talks about the related work and datasets that are gathered for the English language. In section 3, we briefly talk about FarBase. In section 4, we demonstrate the workflow of our method for generating complex questions over Farsbase. Section 5 Some characteristics of the dataset are presented. In section 6, we conclude the paper and talk about future works.

## II. Related Work

As machine learning techniques showed more promising results in the task of question answering, the need for large datasets is increased. Creating such datasets can be done both by crowdsourcing or using templates and rules over raw texts or knowledge graphs. Moreover, simple questions or queries can be used to generate complex questions as well.

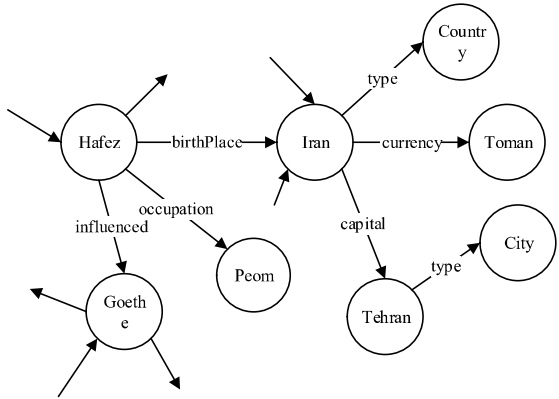

Fig. 1. FarsBase subgraph.

Question answering over knowledge graph concentrates on extracting the answer from the information and knowledge coded in a knowledge graph. It needs training datasets that store the question along with either the piece of knowledge graph containing the answer or the query which extracts the answer from the knowledge graph. WebQuestions and ComplexQuestions (having 2,100 complex questions) gathered questions by using search engine queries. They utilized rule-based methods to ensure the presence of the query's information in the Freebase. ComplexWebQuestions (having 34,689 complex question-answer pairs) is created based on WebQuestionsSP [14] by adding constraints to the SPARQL of its simple questions to create complex questions. ComQA [1] is created over the WikiAnswers community QA platform with crowdsourcing efforts. LC-Quad 2.0 contains 30,000 complex questions and answers with paraphrases and corresponding SPARQL queries over Wikidata [15] and DBpedia that is created using templates. QALD[1] is a series of campaigns for evaluating question answering over linked-data that has complex questions with their answers. Each dataset in QALD is generated by different techniques using crowdsourcing, templates, or rules.

Systems using datasets that are created based on texts mainly focus on answering questions that can be answered from raw text. TREC-QA[2] and CLEF-QA[3] are well-known benchmark dataset over the text that have been created using crowdsourcing.

## III. FARSBASE

FarsBase is a Persian knowledge graph that its data is extracted from Persian version of Wikipedia [12]. Similar to other English knowledge graphs, FarsBase is a collection of subject-relation-object triples (*e1, r, e2*), where *e1* and *e2* are the entities (e.g., *Hafez* or *Iran*) and *r* is a relation/predicate that connects two related entities like *bithPlace*. A triple is also called a fact. There is an example of a subgraph of knowledge graph in Fig. 1.

There are several differences between FarsBase and Freebase (a well-known knowledge graph in English). There are non-real-world entities called CVT [4] nodes in Freebase which are not available in FarsBase. CVTs are only used to collect multiple relations of an event. Another difference is that Freebase entities have unique ids that are used in SPARQL instead of the exact string label of the entity. In FarsBase, there are no ids for entities and no CVT nodes. The absence of CVT nodes has made it easier to generate questions in our approach. While the lack of unique ids makes it hard to find an entity in the knowledge graph as adding a single space or unnormalized character changes the entity's string.

One of the important characteristics of the knowledge graphs is the existence of data type indices. These indices are on the value of typed literals that are stored in knowledge graphs such as xsd:double, xsd:integer, xsd:datetime, etc. Ordinal operations are highly dependent on these data type indices. In FarsBase, some indices of entities are missing and they are stored as a string. This can cause problems in generating complex questions which we will discuss in section IV.

Before explaining our method for generating questions, first we need to explain the definition of a simple and complex question with respect to knowledge graphs. A single-hop (simple) question needs a single fact to arrive at the answer. However, a Multi-hop (complex) question needs multiple edges of KG to find the answer. The complexity of a question is both having an n-hop chain in it or having constraints to limit their answers.

## IV. DATASET GENERATION METHODOLOGY

This section describes the methods we use to generate complex questions. The dataset is produced based on FarsBase by using some rules and templates. Since the knowledge graph is in the form of a graph, we use the nodes (entities) and the edges (their relations) to create questions. In FarsBase each entity has a type such as city, country, film, etc. We collect all the entities of 17 types that are shown in Table I in different files.

TABLE I
TYPES OF ENTITIES THAT ARE USED IN THE DATASET

| Actor   | Director      | Scientist    | Planet     |
|---------|---------------|--------------|------------|
| Athlete | River         | SoccerPlayer | University |
| City    | MusicalArtist | Philosopher  | -          |
| Country | Sea           | Ship         | -          |
| Company | Book          | Film         | -          |

There are three main categories of complex questions in *PeCoQ*: 1) Multi-hop and Multi-entity questions. 2) Ordinal

---

[1] http://qald.aksw.org/
[2] https://trec.nist.gov/data/qamain.html
[3] http://www.clef-initiative.eu/track/qaclef
[4] Compound value type

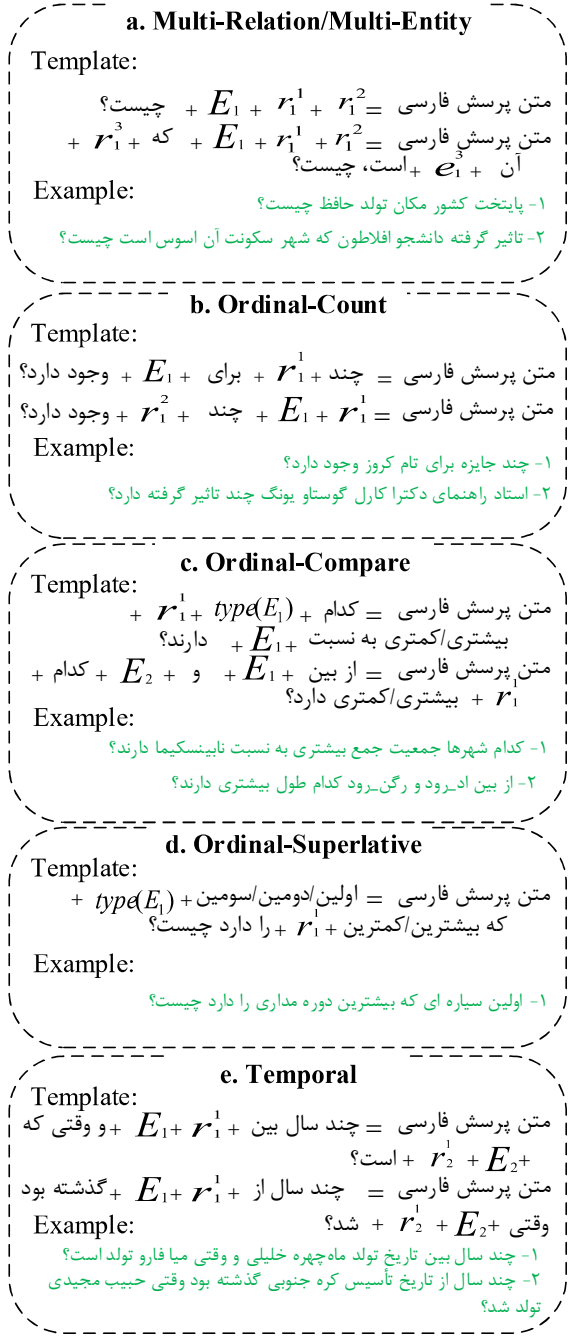

Fig. 2. Example of all categories of complex questions with Persian templates.

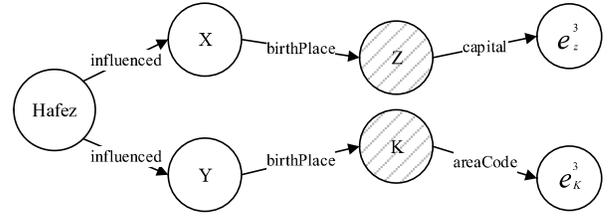

Fig. 3. An example of same hops (*influenced-birthPlace*) with different objects (*Z* and *K*).

### A. Multi-Hop and Multi-Entity

A question that has multiple hops of relations is called a multi-hop question. In this work, we go at most two hops through the graph. For example in Fig. 1 the 2-hop neighboring entities of *Hafez* are: *Country* (birthPlace-type), *Toman* (birthPlace-currency), and *Tehran* (birthPlace-capital). To generate 2-hop questions, we start at a target node that is selected randomly from the entity files. All triples, which have the target entity as its subject, are extracted from FarsBase. To go one-hop further, the objects of retrieved triples are considered as subject and again we extract all triples with the new subjects. Question text is generated with the template of the form (first template in part *a* in Fig. 2):

$$question\ text = what\ is + r_1^2 + r_1^1 + E_1\ ?$$

$Variable_j^i$ shows the entity or relation from *ith* hop of the *jth* target entity. $r_1^2$ and $r_1^1$ are the relations[5] in the second hop and in the first hop, respectively. $E_1$ is the target entity that we start our transition. The answer is the object entities of the second hop. For example, if we start at node *Hafez* ($E_1$) in Fig. 1 and we go 2-hops further a question that can be generated is shown in Fig. 2a.

During the transition, a relation ($r_1^1$ such as *influenced* in Fig. 3) may have more than one object. These objects may have the same relations ($r_1^2$ such as *birthPlace* in Fig. 3). This will cause producing a question text with different answers. To produce multi-entity questions, we extract all triples of this kind of objects (*Z* and *K* in Fig. 3) and then randomly choose a different fact for each of them to use it as the second entity in the question (e.g., *Z-capital*($r_{Z,1}^3$)-$e_Z^3$ and *K-areaCode*($r_{K,1}^3$)-$e_K^3$). This type of question is generated by a template of the form (second template in part *a* in Fig. 2):

$$Q_{text} = what\ is + r_1^2 + r_1^1 + E_1 + where + its + \\ r_1^3 + is + e_1^3?$$

Some of the Persian templates and examples of this type of questions are shown in Fig. 2. According to Fig. 2, some questions text generated by the patterns may not be fluent in Persian. Hence they are reviewed by linguists to correct them so that the final questions be accurate.

questions such as aggregation, superior, and comparative. 3) Temporal questions.

To begin the transition, first an entity must be chosen as the start node. We select 50 entities randomly for each type and start the transition to create different types of complex questions, which we will discuss in the following sections.

---
[5]We translate English relations into Persian

TABLE II
TOTAL COUNT OF DIFFERENT TYPE OF COMPLEXITY

|  | Type of Complexity | | | | |
| --- | --- | --- | --- | --- | --- |
|  | Multi-Relation/Multi-Entity | Aggregation | Superior | Comparative | Implicit Temporal |
| Train | 3893 | 517 | 48 | 2100 | 1442 |
| Dev | 487 | 64 | 6 | 263 | 180 |
| Test | 487 | 65 | 7 | 260 | 181 |
| Total amount | **4,867** | **647** | **61** | **2,622** | **1,803** |
| Percentage | **49%** | **6.4%** | **0.6%** | **26%** | **18%** |

## B. Ordinal Constraint

Some complex questions require applying mathematical operations on the data to find the answers. Aggregation constraint is one of the operations that a question in this category may need when it contains the enumeration phrases such as *how many*, *count of*, etc. This type of complex questions is generated when a relation has more than one object either in the first hop or in the second hop. In the first hop starting in $E$ in Fig. 3 we have 2 entities that are connected to $E$ by the same relation. The templates is in the form of (Fig. 2b):

$$Q_{text} = how\ many + r_1^1 + does + E_1 + have?$$
$$Q_{text} = how\ many + does + r_1^1 + E_1 + have + r_1^2?$$

The superlative constraint is another operation that a complex question may need. To generate such questions, first we save all the numeric questions (e.g. *totalArea* or *population*). For each relation, a SPARQL query is run over a specific type of entity to get the superior (first, second, or third) entity. The template is in the form of (Fig. 2d):

$$Q_{text} = what\ is\ the\ first + type(E_1)$$
$$+ that\ have\ the\ most + r_1^1?$$

Comparative constraint is an operation between a relation of two entities or one entity with others. Two entities that have the same numeric relation are randomly selected. The question text is generated like table 2. For one entity comparative, we randomly choose an entity with a random numeric relation and compare it to other entities of its type with that relation. The template is the form of (Fig. 2c):

$$Q_{text} = which + type(E_1) + has\ more/less +$$
$$r_1^1 + than + E_1?$$
$$Q_{text} = between + E_1\ and\ E_2 + which\ has$$
$$more/less + r_1^1(r_2^1)?$$

Since some data type indexes of the objects are string and in different units (meter, kilometer, etc.) in FarsBase, for ordinal-compare questions, we save the target entities (in the question) answers' values separately instead of entities which are the real answers.

## C. Temporal Constraint

There are two types of temporal constraints: *Explicit* and *Implicit*. Questions in the first category contain explicit temporal expressions such as *2012*. On the other hand, questions in the second category contain implicit temporal expressions such as *during the Word War* II.

Explicit temporal questions are generated as the second entity ($e_1^3$), as we described in subsection A. Facts that have time relation like *birthDate* or *deathDate* are chosen to create explicit temporal questions as the second entity.

Implicit temporal questions are generated as the distance between the time of happening between two entities. The template is the form of (Fig. 2e):

$$Q_{text} = how\ many\ years\ passed\ between + r_1^1 + E_1 +$$
$$and + E_2 + r_2^1?$$

Since some time objects do not have numerical or temporal type indices and are stored as a string, we only store the answer of each entity, not the exact time interval.

## V. DATASET CHARACTERISTICS AND EVALUATION

In this section, we talk about some statistics of our dataset. Almost 127,000 complex questions were generated which we select 10,000 questions that were produced correctly as the rest of the generated questions had the following problems: 1) Some multi-relation questions did not have any meaning as the information was incorrect in the knowledge graph such as *What is PostalCode of BirthDate of Britney Spears?*. In Farsbase, for the triple $< BritneySpears, BirthDate, x >$, there is an entity of type city located at x. 2) Ordinal patterns produced some ordinal questions that do not have numeric relations in them such as *what is the first country that has the most CapitalCity?*. 3) There were duplicate questions that were caused during the transition on target entities having more than one object for the same relations, such as the example in Fig. 3 and also for superlative questions that only get the target entity's type.

These questions were given to linguists to write two paraphrases and also correct questions error, which were about 0.2% of the dataset. This dataset contains over 456 unique relations and 3,054 unique entities. We split the dataset into three parts: training set (80% of the dataset), development set (10% of the dataset), and test set (10% of the dataset). There are 5 types of complexities that their occurrences are shown in Table II. Simple questions do not exist in this dataset. We release this dataset in JSON format. Some samples of PeCoQ are shown in Fig. 4.

## VI. CONCLUSION AND FUTURE WORK

We introduced a large dataset over FarsBase that contains different types of complex questions with corresponding SPARQL queries, paraphrases, and answers. The dataset is generated semi-automatically by graph transition on FarsBase. The machine-generated questions are given to linguists to write two paraphrasings that can help better training for machine learning techniques.

```
{
  "MachineGenerated": "ملیت تاثیر گذاشته شهاب‌الدین_یحیی_سهروردی که نویسنده فتوحات_مکیه است، چیست؟",
  "Question": [
    "ملیت کسی که شهاب‌الدین یحیی سهروردی بر او تاثیر گذاشته و نویسنده فتوحات مکیه است، چیست؟",
    "کسی که شهاب‌الدین یحیی سهروردی بر او تاثیر گذاشته و فتوحات مکیه را نوشته است، اهل کجاست؟"
  ],
  "QuestionType": "MultiRelationQuestions",
  "Relations": [
    "influenced",
    "nationality",
    "author"
  ],
  "Entities": [
    "شهاب‌الدین یحیی سهروردی",
    "فتوحات مکیه"
  ],
  "SPARQL": "select distinct ?x where { <http://fkg.iust.ac.ir/resource/شهاب‌الدین_یحیی_سهروردی> fkgo:influenced ?o. ?o fkgo:nationality ?x. <http://fkg.iust.ac.ir/resource/فتوحات_مکیه> fkgo:author ?o. }",
  "PreAnswers": [],
  "Answers": [
    "عرب",
    "اندلس"
  ]
}

{
  "MachineGenerated": "کدام دانشمندان تاریخ تولد بیشتری به نسبت مل_تامپسون دارند؟",
  "Question": [
    "چه دانشمندانی دیرتر از مل تامپسون متولد شده‌اند؟",
    "کدام دانشمندان پس از مل تامپسون به دنیا آمده‌اند؟"
  ],
  "QuestionType": "Ordinal_Compare",
  "Relations": [
    "birthDate"
  ],
  "Entities": [
    "مل تامپسون"
  ],
  "SPARQL": "select ?s where { ?s fkgo:birthDate ?o2. fkgr:مل_تامپسون fkgo:birthDate ?o1. ?s rdf:type fkgo:Scientist. filter(?o2 > ?o1) }",
  "PreAnswers": [
    "۱۹۴۶_میلادی"
  ],
  "Answers": []
}
```

Fig. 4. Samples of questions in JSON format.

Through our discussion, complexities are bounded and more complex questions can be created on this dataset using their SPARQL or machine-generated text. These include (1) going further hops rather than 2-hops in this dataset, (2) combining different constraints rather than having only a single constraint in a question, (3) solving FarsBase shortcomings or postprocessing the answers of ordinal and temporal questions. These steps are assumed as the future work of this paper.